\documentclass[12pt]{article}

\usepackage{amssymb,amsmath,amsfonts,latexsym,graphicx}
\usepackage[numbers]{natbib}

\begin{document}

\begin{center}
\Large \bf Capsule Vision Challenge 2024: Multi-Class Abnormality Classification for Video Capsule Endoscopy \rm

\vspace{1cm}


\large  Aakarsh Bansal$\,^a$,  \large Bhuvanesh Singla$\,^a$, \large  Raajan Rajesh Wankhade$\,^a$, \\\large  Dr. Nagamma Patil$\,^a$

\vspace{0.5cm}

\normalsize


$^a$ Dept. of Information Technology, National Institute of Technology, Karnataka

\vspace{5mm}


Email: {\tt raajanrajeshwankhade@gmail.com}

\vspace{1cm}

\end{center}

\abstract{
This study presents an approach to developing a model for classifying abnormalities in video capsule endoscopy (VCE) frames. Given the challenges of data imbalance, we implemented a tiered augmentation strategy using the albumentations library to enhance minority class representation. Additionally, we addressed learning complexities by progressively structuring training tasks, allowing the model to differentiate between normal and abnormal cases and then gradually adding more specific classes based on data availability. Our pipeline, developed in PyTorch, employs a flexible architecture enabling seamless adjustments to classification complexity. We tested our approach using ResNet50 and a custom ViT-CNN hybrid model, with training conducted on the Kaggle platform. This work demonstrates a scalable approach to abnormality classification in VCE. 

}

\section{Introduction}\label{sec1}
\textbf{Video capsule endoscopy (VCE)} is a minimally invasive diagnostic technique that involves swallowing a small, camera-equipped capsule to capture images of the gastrointestinal (GI) tract as it passes through. This procedure allows for detailed visualization of the small intestine, an area challenging to reach with traditional endoscopic methods. 

VCE has proven useful for identifying various GI abnormalities, including bleeding, erosions, angioectasia, polyps, and foreign bodies, among others. Despite its benefits, VCE produces a vast amount of image data, which requires time-consuming analysis by medical professionals to detect abnormalities effectively. Given these challenges, there is growing interest in developing AI-based systems to automatically classify abnormalities in VCE images, thereby aiding clinicians in their diagnostic efforts.

The huge amount of data requires careful labeling by medical experts which adds to the cost. There is also a concern of privacy while sharing these images. All these factors lead to formation of a dataset which is imbalanced towards certain classes and rarity of certain diseases increases the dataset's skewness which makes it difficult to be used for training. To address this, we applied a systematic approach to data augmentation using the albumentations library \cite{albumentations}. Augmentation techniques were grouped into three levels, Heavy, Medium, and Light, based on their intensity. These augmentations included transformations such as horizontal and vertical flips, rotations, color jittering, and Gaussian blur. By expanding the diversity of minority-class images through this approach, we aimed to balance the training data distribution and enhance model performance on underrepresented classes. 

Additionally, we structured the learning process to begin with a straightforward task, classifying Normal vs. Abnormal cases. We gradually increased task complexity by adding specific abnormal classes one at a time. This progressive approach allowed the model to build a strong foundational understanding before tackling more complex, data-scarce classes. Experiments with both ResNet50 \cite{resnet} and a custom ViT-CNN hybrid architecture were conducted. 
This strategy highlights an adaptable framework for developing AI-based VCE classification models capable of addressing data imbalance and learning challenges, fostering improved generalization across VCE datasets. 
\section{Methods}\label{sec2}
Our methodology mainly revolves around two aspects:
\begin{itemize}
\item Addressing the pre-existing imbalance in the train dataset.
\item Addressing learning issues that the model might face when learning on imbalanced data.
\end{itemize}

\subsection{Addressing Data Imbalance}
The dataset in \citeauthor{Handa2024} was heavily imbalanced, containing 28,663 Normal images for training and only 158 images for the Worms class. All abnormal classes had less than 3,000 images compared to 28,663 for the Normal class.

To address this imbalance, we developed three levels of augmentations adapted from the albumentations library: 
\begin{itemize}
    \item Heavy Augmentations
    \begin{itemize}
        \item HorizontalFlip
        \item VerticalFlip
        \item RandomRotate90
        \item ColorJitter
        \item ShiftScaleRotate
        \item GaussianBlur
    \end{itemize}
    \item Medium Augmentations
    \begin{itemize}
        \item HorizontalFlip
        \item VerticalFlip
        \item RandomRotate90
        \item ColorJitter
    \end{itemize}
    \item Light Augmentations
    \begin{itemize}
        \item HorizontalFlip
        \item VerticalFlip
    \end{itemize}
\end{itemize}

\begin{figure}[h!]
     \centering
     \includegraphics[width=0.75\linewidth]{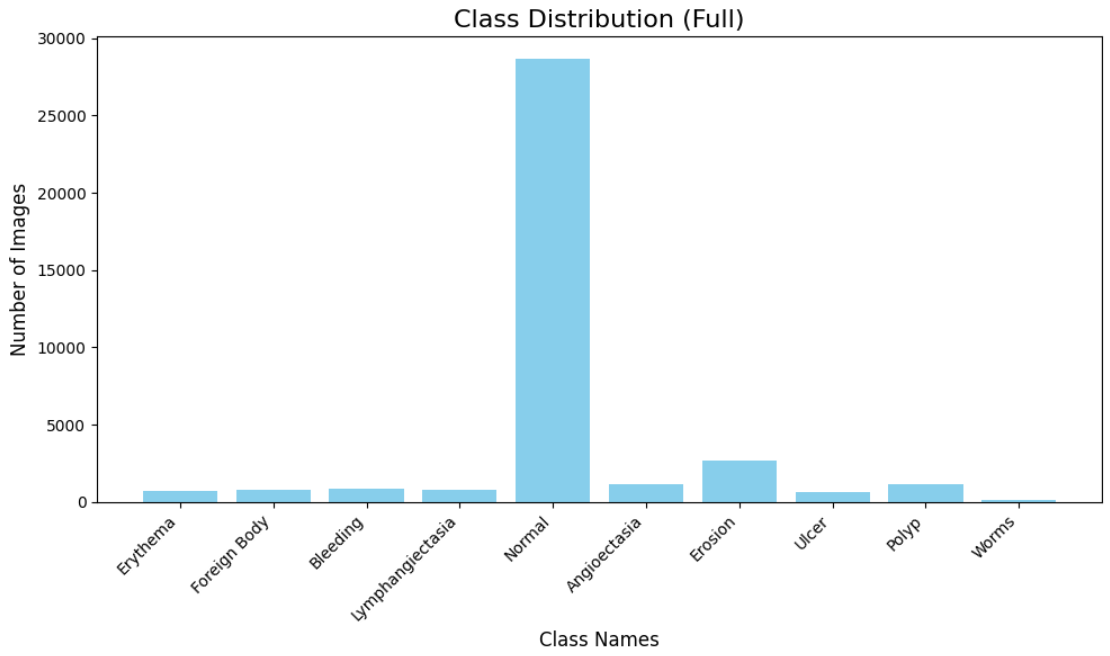}
     \caption{Pre-augmentation Class Distribution (with Normal class)}
     \label{fig:normal_class_inc}
 \end{figure}

Class Distribution in train set before the augmentations are shown in Figure \ref{fig:normal_class_inc}.

\begin{figure}[h!!]
    \centering
    \includegraphics[width=0.75\linewidth]{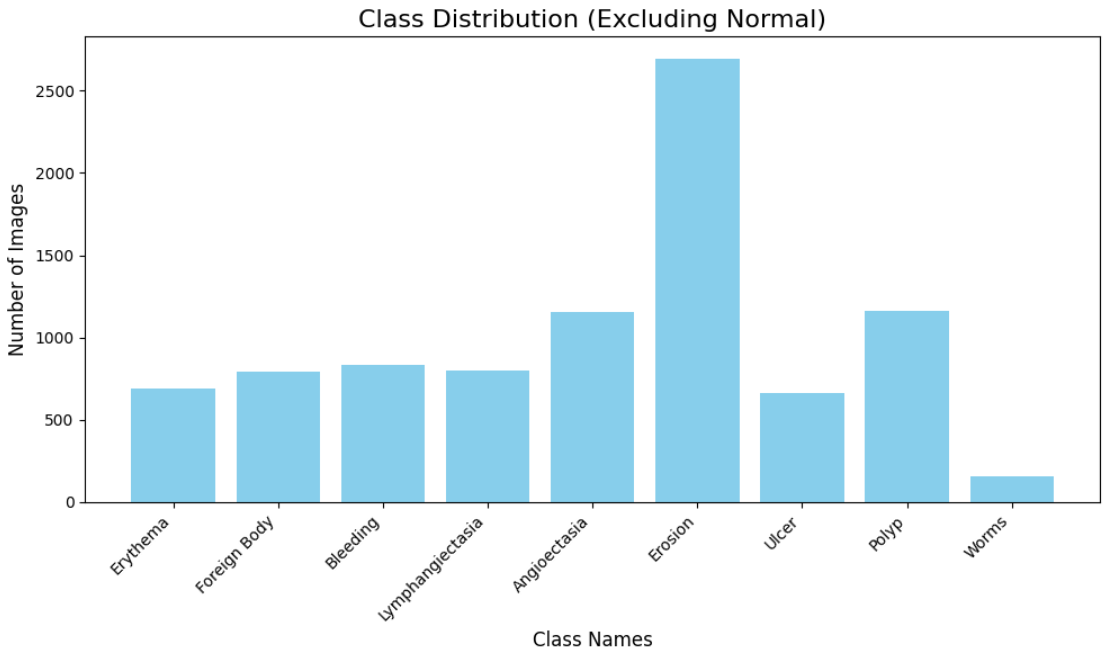}
    \caption{Pre-augmentation Class Distribution (without Normal class)}
    \label{fig:wo_norm}
\end{figure}

Class Distribution in train set before augmentations, excluding the normal class, has been shown in Figure \ref{fig:wo_norm}

\begin{figure}[h!!]
     \centering
     \includegraphics[width=0.75\linewidth]{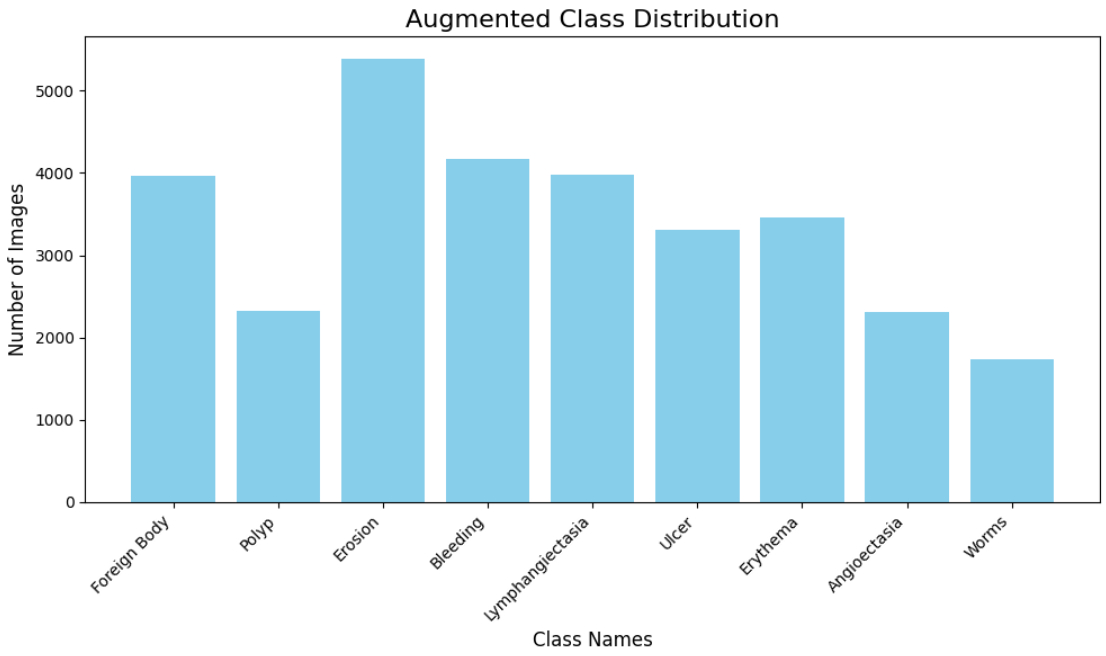}
     \caption{Post-augmentation Class Distribution (without Normal class)}
     \label{fig:augmented_dist}
 \end{figure}

Class Distribution in train set after augmentations, excluding Normal class, has been shown in Figure \ref{fig:augmented_dist}.

\subsection{Addressing Learning Challenges}

When working with an imbalanced dataset, it becomes difficult for the model to learn effectively. To handle this, we used ideas from \cite{curriculum-learning} and adapted them for VCE Image Classification. The main idea is to start with the simplest task, which is to classify Normal vs. Abnormal cases, and then gradually introduce one abnormal class at a time in each step. A similar method is used in \cite{one-at-a-time-ref0}, where a "harder" task is based on how much annotators agree. In our case, however, we define a "harder" task by the number of training images available, with the class with the fewest images being the hardest.

By slowly increasing the complexity of the task, beginning with the easiest, we aim to help the model learn better in the presence of imbalanced data.

\subsection{Final Pipeline and Implementation}
The final pipeline has been implemented in PyTorch. 
We have defined a custom dataloader that is flexible in handling the classes on which the model is being trained. 
This allows us to simply change the classification head of the model, retain the backbone and introduce new, harder classes in the same code.

\begin{figure}
    \centering
    \includegraphics[width=1\linewidth]{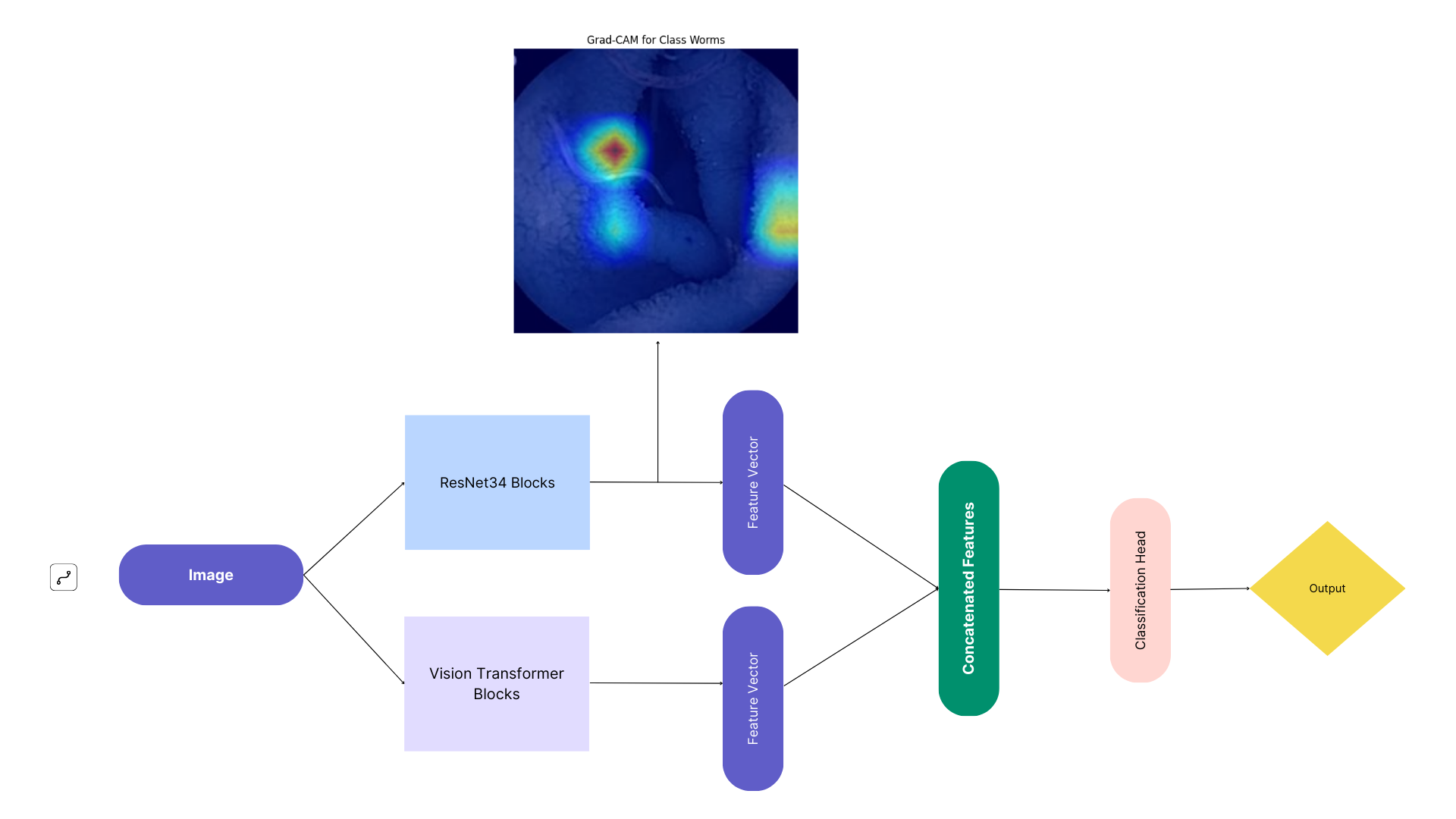}
    \caption{ViT-CNN Architecture and GradCam}
    \label{fig:ViT-CNN}
\end{figure}

We have experimented with the ResNet50 model and a custom ViT-CNN Hybrid (Figure \ref{fig:ViT-CNN}) architecture. The hybrid architecture consists of a ViT branch \cite{visiontransformer} and a ResNet34 branch using which features are extracted. These features are then passed through a classification head to obtain the output.
We have used the Adam \cite{adam} optimizer with a learning rate of 1e-3.
The models were trained on the Kaggle platform using the P100 GPU.

\begin{figure}[h!]
    \centering
    \includegraphics[width=0.5\linewidth]{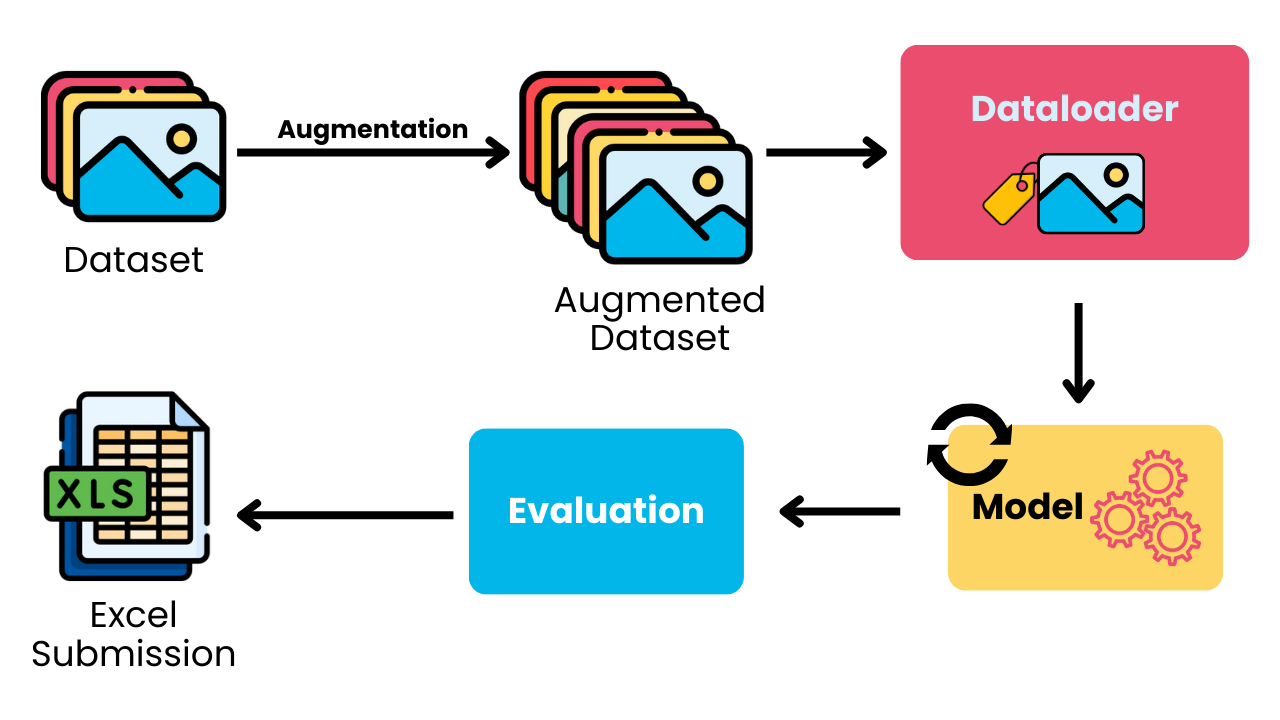}
    \caption{Block diagram of the developed pipeline.}
    \label{fig:blk-diagram}
\end{figure}

\subsection{Explainabiltiy}
Although applications of AI in medical imaging are increasing, it is still mostly represented as a blackbox \cite{xAI}. Hence, it becomes necessary to have some understanding behind the model's decisions. There are several ways of doing this, some methods add explainability from the beginning of training (ante-hoc) while some post training (post-hoc). Since we are using pretrained models with fine tuning, we have implemented a post-hoc technique called Gradient-weighted Class Activation Mapping (GradCAM) \cite{gradCAM}. This technique, as per \citeauthor{xAI} has performed well for medical imaging. Therefore, we applied this technique to the CNN (pretrained ResNet34) branch of our hybrid ViT-CNN model in order to gain insights behind the decisions taken by the model, an example is shown in Figure \ref{fig:ViT-CNN}.

\section{Results}\label{sec3}
The results obtained showed a notable increase in validation accuracy and F1 scores when trained using our training method as compared to direct training.

We have obtained results for standard training using ResNet50 and ViTCNN model on the augmented dataset and compare it with our training procedure on the augmented dataset.
\subsection{Achieved results on the validation dataset}
We can see that the inclusion of our Methodology described in sections 2.1 and sections 2.2 increases the overall performance of the model. A comparative analysis has been shown in  Table \ref{tab:results table}.



\begin{table*}[h!]
\centering
\caption{Validation results and comparison to the baseline methods reported by the organizing team of Capsule Vision 2024 challenge.}

\label{tab:results table}
\begin{tabular}{ccccc}
\hline
\textbf{Method}              & \textbf{\begin{tabular}[c]{@{}c@{}}Avg. \\ ACC\end{tabular}} & \textbf{\begin{tabular}[c]{@{}c@{}}Avg.  \\ Precision\end{tabular}} & \textbf{\begin{tabular}[c]{@{}c@{}}Avg. \\ Recall\end{tabular}} & \textbf{\begin{tabular}[c]{@{}c@{}}Avg. \\ F1-score\end{tabular}} \\ \hline
\textbf{ResNet50 (baseline)}    &         76\%&           0.78&         0.76&          0.76\\
\textbf{SVM (baseline)} &         82\%&           0.81&         0.82&             0.78\\ \hline
\textbf{ResNet50} (Our Method)&         89.57\%&           0.893&         0.895&             0.894\\ \hline
\textbf{ViT-CNN Hybrid} (Our Method)&         \textbf{89.79\%}&           \textbf{0.909}&         \textbf{0.897}&             \textbf{0.902}\\ \hline
\end{tabular}
\end{table*}

Note: All averages here are weighted averages.

\section{Discussion}\label{sec4}

Our methodology tries to structure the learning process such that the model starts with an easy tasks and the  complexity of the classification tasks increase gradually.
This method proved effective as is seen in  Table\ref{tab:results table} where we can see a notable increase in the F1 scores of the model. This signifies that the model has become better at handling minority classes (the harder tasks in our case). 

\section{Conclusion}\label{sec5}
In summary, the approach described here successfully tackles the challenges of classifying imbalanced image data for the Capsule Vision 2024 challenge. By starting with an easier task, separating Normal from Abnormal, and gradually adding more specific categories, the model learns to handle the complexity of the data step by step. This strategy, inspired by curriculum learning, helps the model adjust to the difficulty of each class, especially those with fewer examples. Overall, this method shows promise for improving results across all classes, as seen in the results compared to baseline models.
\section{Acknowledgments}\label{sec6}
As participants in the Capsule Vision 2024 Challenge, we fully comply with the competition's rules as outlined in \cite{handa2024capsule}. Our AI model development is based exclusively on the datasets provided in the official release in \cite{Handa2024}.

\bibliographystyle{unsrtnat}
\bibliography{main}

\end{document}